%% file: main.tex
\documentclass[12pt]{article} 
\usepackage{amsmath,amsthm,amssymb,bm,mathrsfs}
\usepackage{graphicx,subfigure}
\usepackage{caption,enumerate,listings,setspace}
\usepackage[colorlinks,linkcolor=red,anchorcolor=blue,citecolor=blue]{hyperref}
\usepackage[margin=1in]{geometry}
\usepackage[title]{appendix}
\usepackage{enumitem}
\theoremstyle{plain}
\usepackage{booktabs}
\usepackage[numbers]{natbib}
\usepackage{array}
\usepackage{algorithm}
\usepackage{algpseudocode}
\usepackage{makecell}
\usepackage{tikz}
\usepackage{bbding}
\usepackage{cancel}
\usepackage{amsthm}
\usepackage{graphicx}
\usepackage{wrapfig}
\usepackage{caption}
\theoremstyle{definition}

\input{def-smile}

\input{front}

\title{\textbf{GReaTER: Generate Realistic Tabular data after data Enhancement and Reduction}} 
\date{March 20, 2025}

\author
{ Tung Sum Thomas Kwok \thanks{Incoming Ph.D. Student, Department of Statistics and Data Science, UCLA, CA, 90095. Email: tk1018@ucla.edu}, 
Chi-Hua Wang \thanks{Postdoctoral Scholar, Department of Statistics and Data Science,  UCLA, CA, 90095. Email: chihuawang@ucla.edu},
Guang Cheng\thanks{Professor, Department of Statistics and Data Science, UCLA, CA, 90095. Email: guangcheng@ucla.edu}
}

\begin{document} 

\maketitle

\begin{abstract}
Tabular data synthesis involves not only multi-table synthesis but also generating multi-modal data (e.g., strings and categories), which enables diverse knowledge synthesis. However, separating numerical and categorical data has limited the effectiveness of tabular data generation. The GReaT (Generate Realistic Tabular Data) framework uses Large Language Models (LLMs) to encode entire rows, eliminating the need to partition data types. Despite this, the framework's performance is constrained by two issues: (1) tabular data entries lack sufficient semantic meaning, limiting LLM's ability to leverage pre-trained knowledge for in-context learning, and (2) complex multi-table datasets struggle to establish effective relationships for collaboration. To address these, we propose GReaTER (Generate Realistic Tabular Data after data Enhancement and Reduction), which includes: (1) a data semantic enhancement system that improves LLM's understanding of tabular data through mapping, enabling better in-context learning, and (2) a cross-table connecting method to establish efficient relationships across complex tables. Experimental results show that GReaTER outperforms the GReaT framework.
\end{abstract}


\clearpage
\section{Introduction}
Tabular data generation is a form of generative AI focused on creating tabular data, gaining attention in areas like privacy-preserving analytics \cite{privacy} and data augmentation \cite{Mumuni_2024}. It has facilitated applications such as road scene detection \cite{traffic}, disease detection \cite{disease}, and customer transaction analysis \cite{bank}. However, most traditional tabular synthesis methods focus on single-modal data (either all numerical or all categorical), overlooking the prevalence of multi-modal data. To address this, the GReaT framework \cite{borisov2023language} integrates diverse data types and uses minimal transformation to better preserve real-world characteristics, as shown in Fig. \ref{fig:GReaT_implementation_problem}. However, two challenges still remain, making the direct use of GReaT ineffective.\\
\\
\textit{Challenge I} is the lack of a fine-grained categorical data conversion method in textual encoding, which is crucial for LLM data understanding \cite{solatorio2023realtabformer}. The existing GReaT framework learns feature distributions based on historical tabular value occurrences. However,  t cannot distinguish numerical labels used in different features to represent separate categories. By way of illustration, Fig. \ref{fig:GReaT_implementation_problem} shows that multiple '1's are tokenized identically, even though they represent different categories. This ambiguous tokenization further creates false relationships by associating co-occurring numerical labels across unrelated features, as illustrated in the '1's in 'Lunch' and 'Access Device' in Fig. \ref{fig:GReaT_implementation_problem}, misleading the model into assuming an unwarranted dependency. Finally, LLM robustness is hindered by the lack of semantic meaning in numerically labeled data (e.g., '1's without supporting units). Solving these issues is crucial for learning fine-grained patterns with LLM.\\
\\
Challenge II involves the limited handling of multi-table data in LLM-based synthesizers. Flattening multi-table data for textual encoding often introduces noise that reduces synthetic fidelity. This is because the introduction of multi-table data further exacerbates the occurrence of active / engaged subjects (e.g. users), which leads to decision bias of LLM \cite{derec}. Multi-table synthesizers suffer from this engaged subject bias, struggling to establish effective relationships across complex tables. While the DEREC \cite{derec} pipeline attempts to address this, it has several limitations: (1) It fails to consider all features simultaneously. (2) It lacks modeling consistency, leading to redundant understanding of parental table distributions over different parent-child table modeling. (3) It is computationally inefficient to model two sets of parent-child tables separately.\\
\\
To address the above challenges, we propose a novel GReaTER (Generate Realistic Tabular data after data Enhancement and Reduction) model, which enhances numerical label semantics and models all features from multiple tables in one shot. Specifically, it includes two key components: (1) Challenge I is tackled by improving numerical label semantics through a data semantic enhancement system which helps LLM differentiate and understand data. (2) Challenge II is addressed with a cross-table linkage method using dimensional reduction and feature correlation. In summary, this study makes the following contributions.

\begin{itemize}
    \item To the best of our knowledge, we are the first to investigate how to effectively leverage tabular data semantics to conduct multi-modal and multi-table data synthesis.
    \item The above problems are solved via our proposed GReaTER model, which improves data semantics for label understandability, and alleviates the problem of engaged subject bias in multi-table synthesis.
    \item Extensive experiments on two benchmark synthesizers have been conducted, with results demonstrating better performance of GReaTER over state-of-the-art baselines.
\end{itemize} \vspace{-6mm}
\textcolor{white}{\footnotesize{blank}}\\ \vspace{-2mm}
\\ 
The rest of the paper is organized as follows. Firstly, we review the related work in Section \ref{sec: related-work}. Then, Section \ref{sec: methodology} presents GReaTER to complete the multi-modal and multi-table data generation. Experiments are conducted in Section \ref{sec: experiments}, which is followed by the conclusion and future work in Section \ref{sec: further-discussion}. Preliminary definitions of key terms are included in Appendix \ref{def:contextual-variable-notations} and \ref{def:parent-and-child}. 
\begin{figure*}
    \centering
    \includegraphics[width = 0.95 \linewidth]{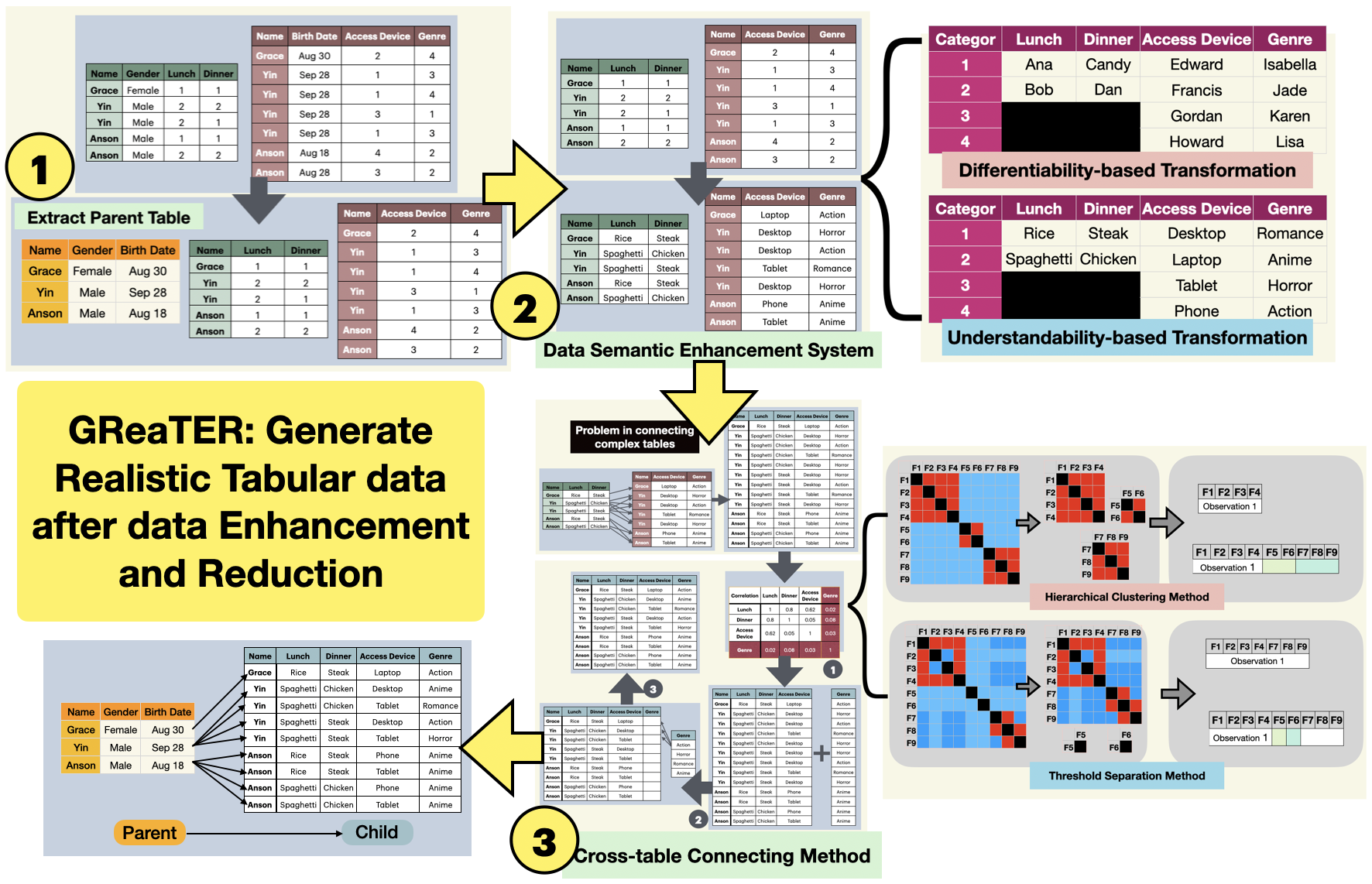}
    \caption{Overview of GReaTER:(1) Extract the parent table for multi-table synthesizer based on contextual variables (2) Improve data semantic level for textual encoder to transform tabular data into semantically meaningful sentence to train LLM. (3) Join the remaining two child tables together while reducing engaged subject bias from direct flattening. }
    \label{fig:greater-overview}
    
\end{figure*}

\section{Related Work} \label{sec: related-work}
\subsubsection*{LLM usage in tabular data synthesis} Tabular data synthesis is essentially a generative artificial intelligence designed specifically to process and generate tabular data. As LLM is trained under internet data rich of multi-modal characteristics, it is proposed to integrate with multi-modal data in tabular data synthesis. GReaT is composed of a textual encoder (i.e., translator to convert tabular row into a sentence), and LLM backbone (i.e., GPT-2 fine-tuned from tabular data) \cite{borisov2023language}. The above method ignores the following details: (1) repeating numerical categories in different features can impair the robustness of model; (2) oversimplified numerical categories do not contain any semantics, which limits the utilization of semantic enhancement using LLM prompts.
\begin{figure} % 'r' for right, 'l' for left
\centering
    %\includegraphics[width = 0.7
    %\linewidth]{Figures/GReaTER_pipeline.png}
    \includegraphics[width = 0.9 \linewidth]{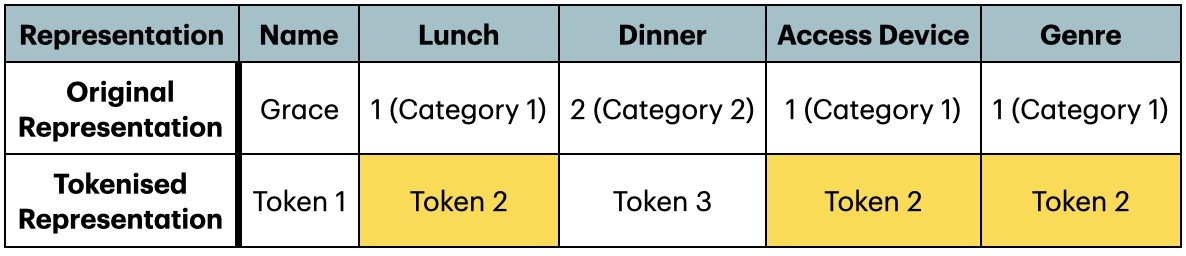}
    \caption{A simple example of GReaT implementation: The row observation is textual-encoded in the form 'Name: Grace, Lunch: 1, Dinner: 2, Access Device: 1, Genre: 1', but the LLM would struggle to differentiate between the repeating '1's and would tokenize the different '1's into the same embeddings. To easily follow our work, we exemplify this implementation throughout the paper. }
\label{fig:GReaT_implementation_problem} 
\end{figure}

\subsubsection*{Complex multi-table synthesis}
Early tabular data synthesis studies focused on single-table generation \cite{tabddpm} \cite{ctgan} \cite{borisov2023language}, but they struggled with multi-table synthesis \cite{derec}. To address this, some studies use hierarchical clustering to simplify multi-table datasets by establishing parent-child relationships \cite{sdv}. This approach improves upon direct flattening methods by linking key tables. However, hierarchical clustering cannot handle different modalities simultaneously without transformation. To overcome this, some propose using textual encoding and LLMs to model parent-child relationships, such as Seq2Seq models for generating child table observations conditioned on parent data \cite{solatorio2023realtabformer}. Yet, these methods rely on the assumption of a strict parent-child relationship, limiting their applicability to realistic datasets. To better process diverse datasets, some studies use table restructuring or bipartite matrix modeling \cite{derec} \cite{alimohammadi2024adapting}, but these methods assume independence between child tables, hindering effective fusion. The inadequacy of existing methods motivates our multi-modal and multi-table fusion goal to simultaneously complete the intra-modal and inter-table interactions in a unified and low-noise way.

\section{Methodology}
\label{sec: methodology}
\subsection{Overview of GReaTER} 
Our proposed model GReaTER, the overview of which is shown in Fig. \ref{fig:greater-overview}, contains two components: (1) a data semantic enhancement system, which conducts precise tabular value interactions by differentiating and improving the label semantics to generate more robust and precise textual-encoded sentence ciphering relevant knowledge of all modalities; (2) a cross-table connecting method, which establishes efficient relationship across complex tables to facilitate cross-table textual encoding with a dimensional reduction mechanism. 

\subsection{Data Semantic Enhancement System} \label{sec: data-semantic-enhacement-system} 
\begin{figure*}
  %\begin{figure}[H]
    \centering
    \includegraphics[width = 0.95 \linewidth]{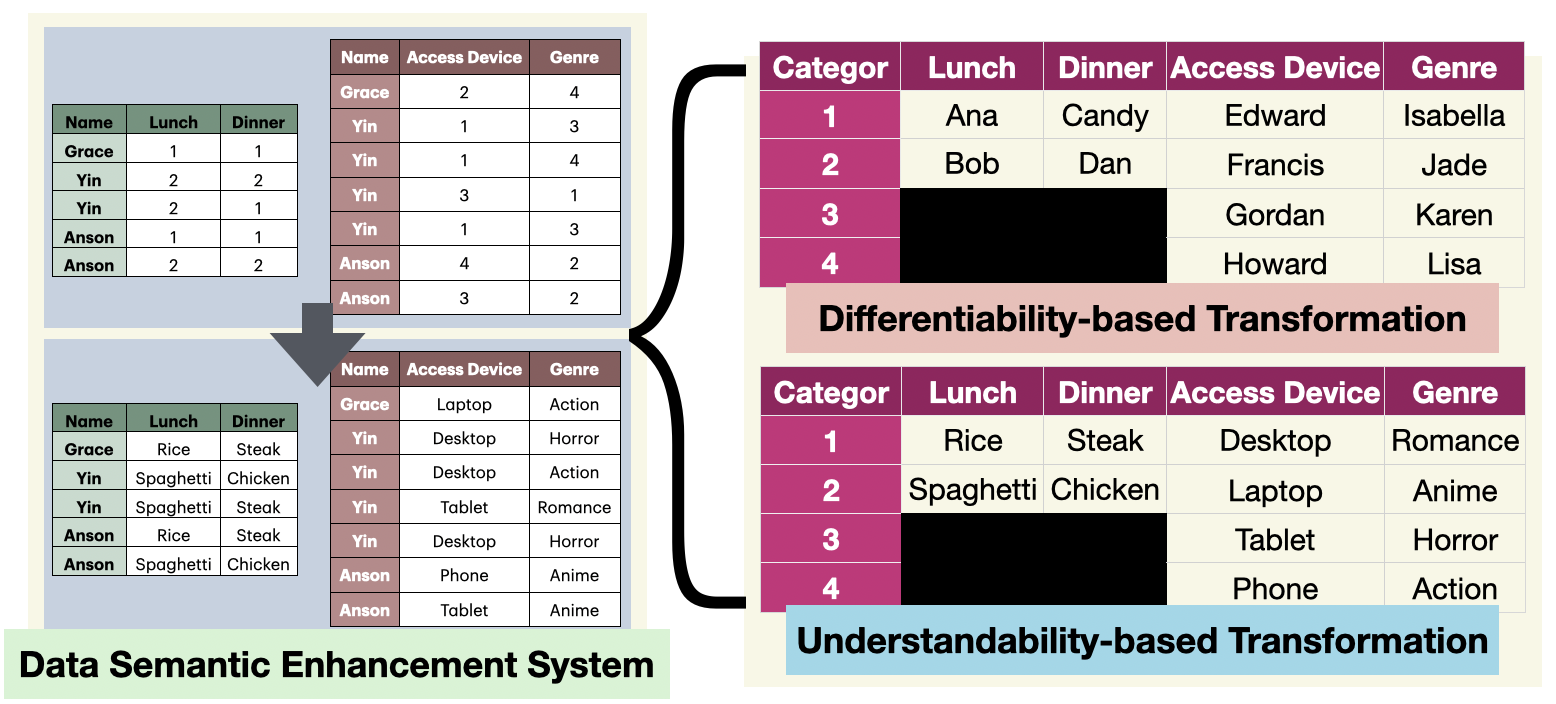}
    \caption{Numerical categories are transformed to unique objects to facilitate LLM in understanding the data. Two transformation mappings are proposed, with one focusing only on differentiabiltiy and the other also taking care of data understandability.}
\label{fig:semantic-enhancement-illus}
\end{figure*}
The data semantic enhancement system addresses ambiguous numerical labels in the GReaT framework, which hinder LLM’s understanding of label-encoded multi-modal data. This system transforms these ambiguous labels by focusing on two key ideas: (1) the impact of intra-table attention on co-occurring numerical categories, and (2) ensuring semantic clarity for LLM in-context learning. It identifies features with co-occurring numerical categories and maps them to semantically distinct and robust words using a mapping system. This approach is novel compared to existing LLM ensembling \cite{xu2024bridginggapdifferentvocabularies} and multi-modal pre-processing methods \cite{sdv}, which are not specific to tabular data generation.\\
\\
The system includes a differentiability-based transformation module and an understandability-based transformation module. It assumes that no mapping is provided by the data source to distinguish label-encoded data. First, selected categorical features are processed through the transformation module, creating a mapping system based on the data. This system is tailored to the desired level of data semantics and aids in subsequent textual encoding for LLM synthesis. The synthetic output is then inversely transformed back into the original data format. Fig. \ref{fig:semantic-enhancement-illus} shows the schematic diagram of this system.

\subsubsection{Differentiability-based Transformation Module} \label{subsub: differentiability-based}
(1) Selected categorical features are processed by the module to determine the number of total categories across all selected features, represented by $n=n_{column 1}+n_{column 2}+...$; (2) An equal amount of unique representations defined as $a_{1}, a_{2},...,a_{n}$ is generated, with each of the category being mapped to one of the representations. While the unique representations do not necessarily relate to the actual semantics, this transformation module seeks to provide minimal but automated differentiability so that there will be no repeating categories anymore in the transformed table. 

\subsubsection{Understandability-based Transformation Module}\label{subsub: understandability-based} To improve understandability of the label encoded categories, we need to create a more precise mapping system that is relatable to the feature name. Precision of word usage has proven useful in enhancing LLM functionality \cite{anonymous2024tabmeta}. However, this semantic addition contains another research paper worth of context, hence is not further discussed in this work. Specifically, all label-encoded entries are mapped to a class expected to occur in the given column. This mapping also guarantees differentiability, providing a stepwise semantic enhancement approach to study LLM synthesizer's sensitivity towards data semantics. To achieve understandability, the label-encoded entries are mapped based on a mapping system designed by data scientists through studying every column. This requires data scientists to have sufficient understanding on the table, and is less scalable. This motivates further work (Sec. \ref{sec: further-discussion}) to automate the module by generating description and most likely categories given column anme and category value using in-context learning of LLM.  \cite{anonymous2024tabmeta}.

\subsubsection{Inverse Mapping System}
\begin{figure}[H]
    \centering
    \includegraphics[width=0.95\linewidth]{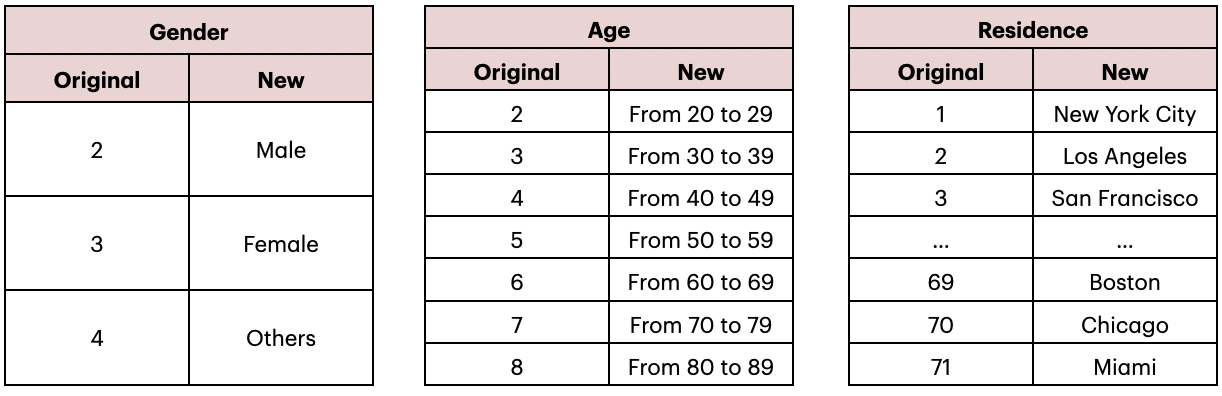}
    \caption{The understandability-based transformation mapping is designed to ensure the value of each category and mapping can be sensibly explained.}
    \label{fig:understandability-based}
\end{figure}

To ensure robustness of the model, we design the inverse mapping system, which is utilized after data synthesis. This is because synthetic product is generated under a synthesizer being trained by the transformed data, whose features are irrelevant in terms of pre-transformed original data format. Specifically, irrelevant semantically enriched categories are irrelevant to the original numerical categories in the multi-modal data, Fig. \ref{fig:understandability-based} illustrates the mapping system so that all the unique subjects within the mapping can be inversely transformed back to the original numerical columns. This motivates the model to always return synthetic data in the same format as the original data. To prevent potential privacy leakage, the mapping system is to be deleted after the data is synthesized, so that malicious users cannot gain access to the mapping system for privacy attacks.

\subsection{Cross-table Connecting Method} \label{sec: cross-table-connecting-method} 
\begin{figure*}[t]
    \centering
    \includegraphics[width=0.95\linewidth]{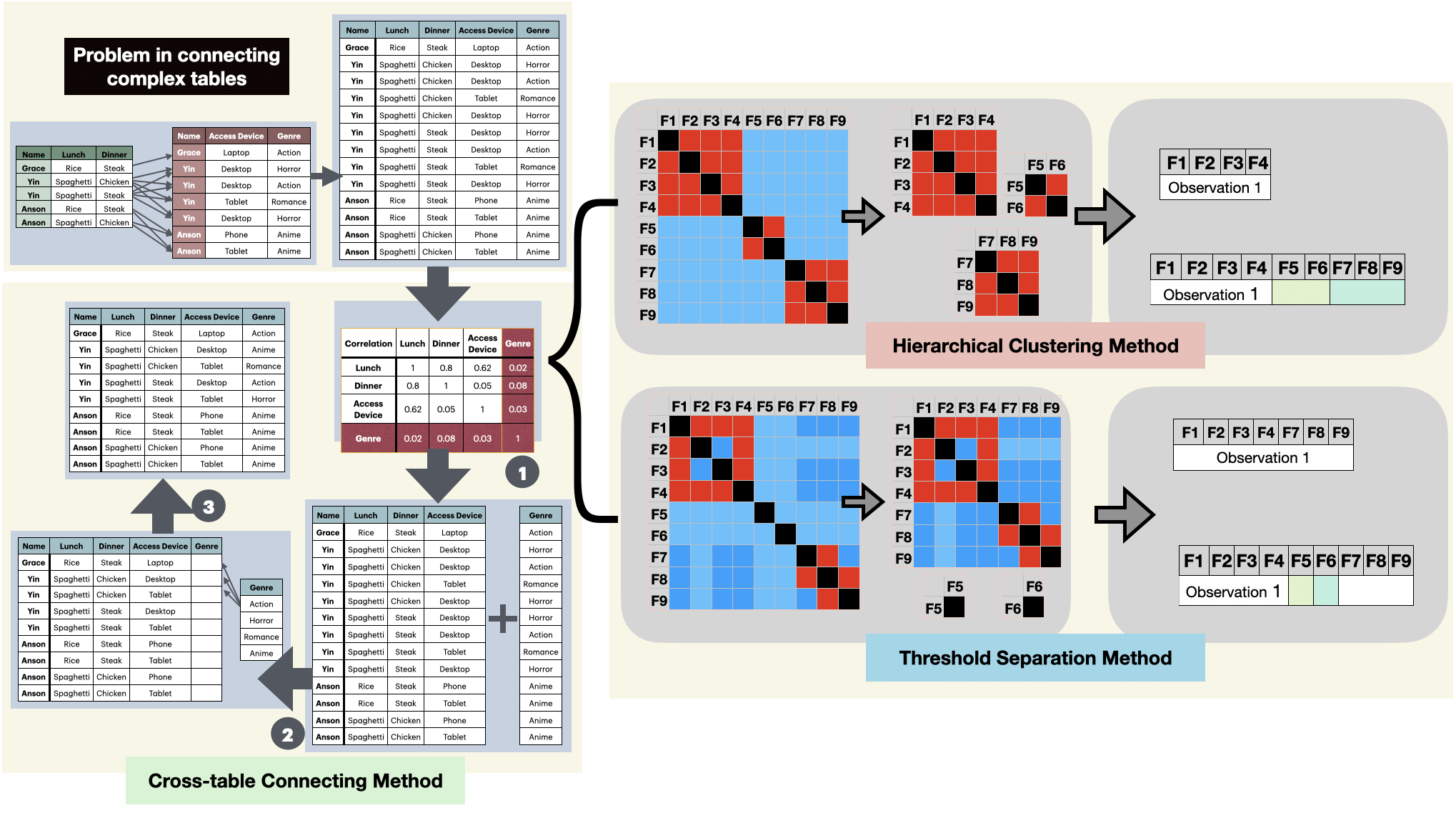}
    \caption{Logic Flow of the Cross-table Connection Method:(0) Flattening two tables creates (0.1) the dimensionality problem: $2\times 5$ table flattened with a $2 \times 7$ table leads to a $13\times 4$ table, and (0.2) engaged subject bias: engaged subject like 'Yin' dominates the distribution with 8 out of 13 observations. (1) Determine columns with low correlation with all other features (Two methods are proposed) (2) Separate these columns from the table (3) Keep only the unique items and append the column back to the table via bootstrap sampling.}
    \label{fig:correlational-dimension-reduction}

\end{figure*}
Existing tabular synthesizers struggle with multi-table scenarios due to the complexity of connecting multiple tables, which can lead to noisy data and engaged subject bias. A simple method, direct flattening, combines two tables, but it (1) creates large, computationally intensive tables and (2) exacerbates engaged subject bias, as more frequent subjects dominate the data, as shown in Fig. \ref{fig:correlational-dimension-reduction} where 'Yin' overpowers less engaged subjects like 'Grace' and 'Anson'. Some approaches restructure tables into a parent-child format \cite{derec}, but this separates child tables (Fig. \ref{fig:many-to-many-combine-with-derec}), ignoring potential interactions between them. To address these issues, we propose a novel cross-table connecting mechanism with two main innovations: (1) a restructuring mechanism that extracts a contextual variable as a parent table and models cross-child-table interactions, and (2) a dimension-reduction method to reduce noise from engaged subject bias.\\
\\
Our GReaTER model upgrades the DEREC framework \cite{derec} by processing both child tables in one round. After creating the parent table, we combine the remaining columns from both child tables, flattening only correlated features and appending independent features via bootstrap sampling. This reduces noise and transforms a many-to-many relationship into a one-to-many structure, using a three-step correlation dimension reduction method: determine independence, reduce dimensions, and append by sampling.

\subsubsection{Determine independence}
To preserve only the informative cross-child-table relationship, we determine independent columns and remove them from the flattened table, e.g. Fig. \ref{fig:correlational-dimension-reduction} shows a correlation matrix to illustrate column independence. In practice, the determination of independent columns can also be done with other tests such as $\chi^{2}$ test and the Fisher's Exact Test, whichever that works best given the dataset. We propose two methods to determine independence, namely a straightforward 'up-and-stay' Threshold Separation method and the Hierarchical Clustering method \cite{cohenaddad2017hierarchicalclusteringobjectivefunctions}. The Threshold Separation method measures the features' correlation coefficient, and treats one feature as independent to the rest if all of its correlation coefficient is less than a certain threshold. The Hierarchical Clustering separates features into different subgroups based on their average pairwise Euclidean distance.
\subsubsection{Reduce dimension}
Hypothesizing that the engaged user bias produces excessive noise, a common approach is to reduce dimension of the dataset. Unlike traditional dimension reduction methods \cite{Jolliffe:1986}, we propose a novel approach that also reduces dimensionality on number of repeating rows after being flattened. Duplicate rows appear when a column is removed, and by removing the duplicate rows, the entire table dimensionality can be reduced. Fig. \ref{fig:correlational-dimension-reduction} shows that duplicate rows can be observed with the 'Genre' column removed ('Yin, Spaghetti, Chicken, Desktop' and 'Yin, Spaghetti, Steak, Desktop'), and removing these rows yields a smaller table. 

\subsubsection{Append by sampling}
The order of values in independent column matters less as they have limited interactions with other features. However, they should still be retained in the table for downstream task. This allows a simple sampling method to append the columns back to the flattened table. To ensure the validity of observations, we propose the bootstrap sampling method where each unique subject contains its own pool, so as to prevent formation of any features combinations that are non-existent in the original data, e.g. Anson in Fig. \ref{fig:correlational-dimension-reduction} only contains 'Anime' in 'Genre', so Anson's pool would only contain 'Anime' for all observation samplings. 

\section{Experiments}
\label{sec: experiments}
\subsection{Experimental Setup}
\subsubsection{Datasets} \label{sec:dataset}
We study the performance of GReaTER on a public dataset, i.e. the Advertisement and Feeds training data from the CTR Prediction - 2022 DIGIX Global AI Challenge \cite{digix_global_ai_challenge_2022}, a multi-table dataset widely adopted in data collaboration studies \cite{derec}. The dataset records data from the advertisement and source domain, with both tables having repeating user IDs as the subject. This dataset has an imbalanced category a clickthrough rate of 1.55\% (3 out of 200 observations are positive) with most features being less informative (correlation ranging at about 0.2). The table issub-grouped based on the task IDs to create more trials, where eight of the subgroups, each with over 750 observations, are chosen to conduct eight rounds of independent trials.

\subsubsection{Dataset Pre-processing}
As most of the features are in categorical form, the Cramer's V coefficient is utilized to quantify the degree of association. In the initial measurement, all features are observed to be highly correlated with each other, so that the flattened table is irreducible. After studying the few columns that are highly correlated to all other features, it is discovered that these features are neither repeating nor categorical. The 'e\_et' refers to a 12-digit number that records the year, month, day and time of the observation, while 'i\_docid' and 'i\_entities' are in the shape of ID addresses. Cramer's V correlation of these features do not have explainable meaning, and may be misleading. Removing these columns gives a less noisy correlation matrix with separable subgroups. 
\begin{figure}[H]
    \centering
  \includegraphics[width=0.9 \linewidth]{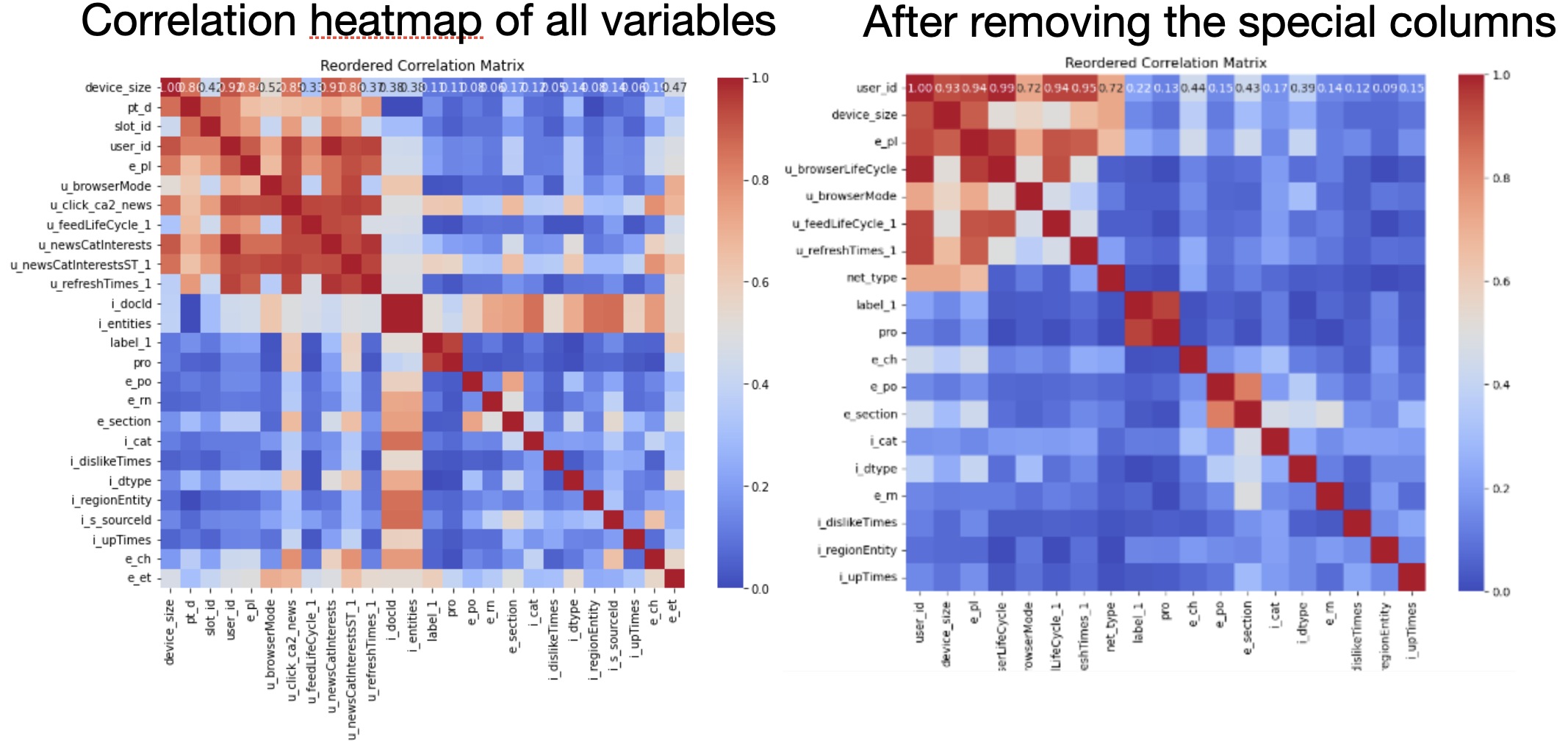}

  \caption{Correlation heatmap before and after columns removal}
    \label{fig:cor_heatmap}
\end{figure}
\subsubsection{Evaluation Protocol}
Synthetic data fidelity are measured in terms of the 'distribution of distribution similarity' metric \cite{derec} to evaluate the performance of GReaTER. For distribution of distribution similarity, we use the following two metrics to quantify similarity over pairwise conditional distribution of original and synthetic data. (1) The p-value of Kolmogorov-Smirnov Goodness-of-Fit Test ("p-value") \cite{kstest} and (2) the Wasserstein Distance ("W-distance") \cite{wdis}, motivated by their simplicity and distribution-free assumption. 

\subsubsection{Hyper-parameters}
 Key training hyperparameters are as follows. Two realtabformer objects created for parent and child tables both have 10 epochs and 5 batches, due to a large dataset size so that the insufficient computing power for default hyperparameters of 1000 epochs with 8 batches.
 
 \subsubsection{Tuning the Data Semantic Enhancement System}
  The differentiability-based transformation (Sec. \ref{subsub: differentiability-based}) converts each category in the selected column into a unique name not appearing in the table, eliminating co-occurring categories. The Python package \texttt{names} \cite{hunner2016names} provides a vast database to generate these unique representations.\\
  \\
  For the understandability-based transformation, categories are mapped to meaningful labels. For example, the gender column's three categories ('2', '3', '4') are interpreted as 'male', 'female', and 'others', with '4' assumed to represent 'others'. Categories '2' and '3' are assigned to 'male' or 'female' without significant performance impact. In the 'Age' column, categories '2' to '8' represent age groups from 20 to 89, based on assumptions about online shopping habits. Finally, the 71 categories in the 'Residence' column, which represent provinces, are mapped to 71 cities in the USA, as most large countries have fewer provinces, and English-speaking cities align with GPT-2's pre-trained knowledge.

 \subsubsection{Tuning the Cross-table Connecting Method}
 Since most pairwise correlation coefficients (Fig. \ref{fig:cor_heatmap}) are very low, both mean and median correlation values are similarly low (0.2 and 0.3), primarily due to the prevalence of values close to zero. As a result, if a feature has pairwise correlation coefficients with all other features below these low mean and median values, it can be considered independent. Therefore, we set the mean and median as the threshold for the Threshold Separation Method. The Hierarchical Clustering method then separates features based on their pairwise average Euclidean distance. 
 
\subsection{Baselines} 
To investigate the performance of GReaTER in multi-modal scenarios, two child table pre-processing methods are compared: 1) direct flattening of the two child tables; 2) conducting two separate rounds of parent-child table synthesis on each child tables \cite{derec}. 

\subsection{Performance Comparisons} 
\begin{figure}[H]
    \centering
    \includegraphics[width=0.95\linewidth]{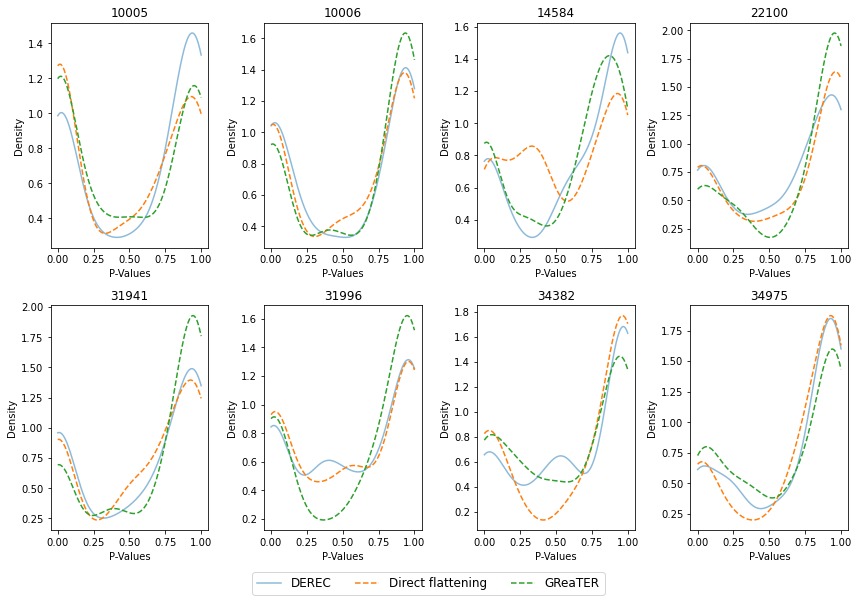}
    \caption{The GReaTER (in green dotted line) generally shows a heavier right tail than the two benchmarks, indicating improved overall synthetic fidelity.}
    \label{fig:overall_results}
\end{figure}

Fig. \ref{fig:overall_results} shows a net improvement in synthetic data fidelity after implementing the GReaTER model over the DEREC benchmark that treats two child tables independently \cite{derec}. Fig. \ref{fig:overall_results} shows that direct flattening the child tables almost always guarantees degraded synthetic fidelity compared to treating them completely independent, which is hypothesized to be noise impact from engaged user bias. To further understand the effect of each component in GReaTER that leads to its outperformance over DEREC, studies regarding each component with multiple setups have been conducted. 
\vspace{-6mm}
\textcolor{white}{\footnotesize{blank}}\\ \vspace{-2mm}
\\ 
Specifically, the Data Semantic Enhancement system is shown helpful in improving fidelity, with a slight improvement after employing the more advanced understandability-based transformation method. The implementation of Cross-table Connecting Method also allows the synthesizer to generate with improved fidelity than DEREC. This shows that there are a considerable degree of correlation between the two child tables, hence implying the value of data collaboration and combined data analysis in producing additional information. 

\subsection{Studies for Data Semantics} 
\begin{figure}[t]
    \centering
    \includegraphics[width=0.95\linewidth]{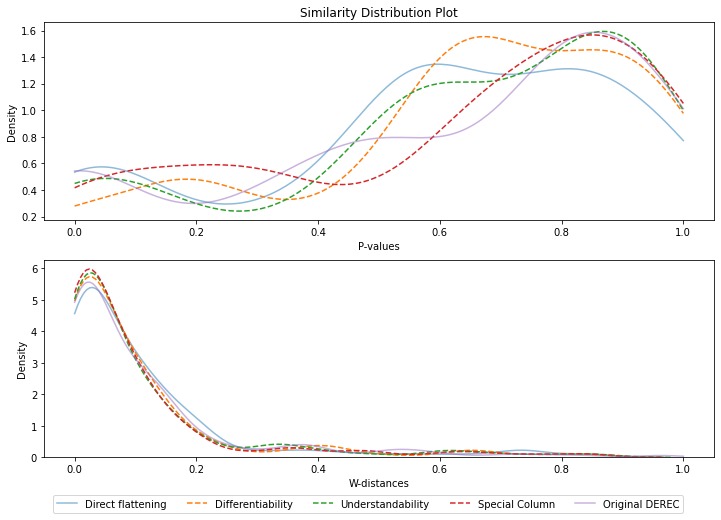}
    \caption{The implementation of GReaTER leads to a heaver p-value distribution on the right, indicating more statistical similar column pairs and hence higher fidelity. This implies LLM performance can be improved by increasing semantic level of the input data. Furthermore, comparing between GReaTER in different semantic setups, we observe further fidelity improvement in using the understandability-based transformation over the differentiability-based transformation. This indicates in-context learning of LLM when given with more precise categories, despite using a rather out-dated GPT2 as the LLM backbone.}
    \label{fig:cat_map_results}
\end{figure}
\subsubsection{Impact of precision of tabular categories} The transformation of categorical variables is crucial for synthetic data fidelity. Fig. \ref{fig:cat_map_results} shows both transformation modules improve fidelity compared to no mapping, confirming that numerical labels across columns confuse LLMs during fine-tuning. Additionally, the understandability-based transformation slightly outperforms the other method, indicating the importance of precise categories in tabular data. This supports the hypothesis that LLMs use these categories and pre-trained knowledge for in-context learning. While GPT-2 shows some in-context learning traits, its ability to capture complex data relationships is limited compared to newer models like GPT-3, which is 100 times larger and more robust, suggesting that advanced LLMs could further improve performance (Sec. \ref{sec: further-discussion}).

\subsubsection{Semantic improvement on dataset-specific columns} \label{subsub: special-transform} The Data Semantic Enhancement Method encourages experimenting with transformations to uncover additional patterns. In the selected dataset, four columns contain values like 20\texttt{\^{}}35\texttt{\^{}}42\texttt{\^{}}15\texttt{\^{}}5, representing product categories of user interest or disinterest. Replacing '\texttt{\^{}}' with 'and' makes the data more natural language-like, improving fidelity. This likely results from LLMs being pre-trained on text, where '\texttt{\^{}}' is rarely used as 'and'. Although this transformation is data-specific and primarily for testing, it supports the idea that providing LLMs with data in more common natural language formats enhances their understanding of tabular data. 
 
\subsection{Studies for Cross-table Connecting Method}
\subsubsection{Necessity to consider cross-child table relationship}
\begin{figure}[t]
    \centering
    \includegraphics[width = 0.95 \linewidth]{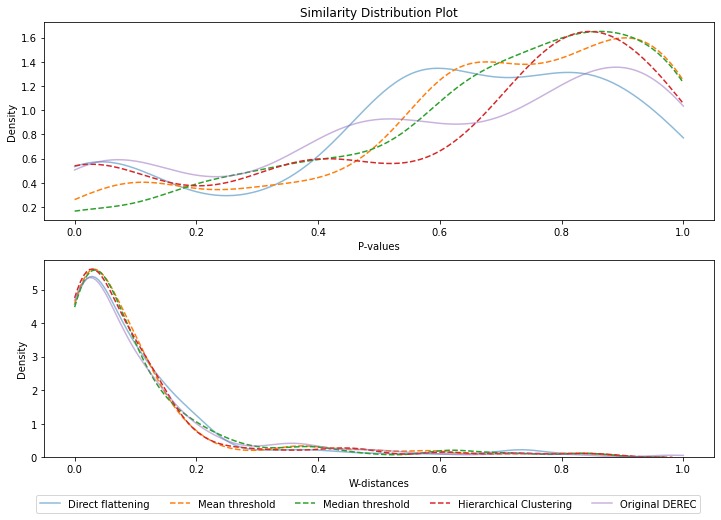}
    \caption{When comparing specifically on different setups to connect multiple tables, it is observed that direct flattening performs the worst. This supports our claim in Sec. \ref{sec: cross-table-connecting-method} when excessive engaged user bias would distort the occurrence distribution. The three Cross-table Connecting cases show similar graphical outperformance, especially for the two threshold setups. It is observed that the Threshold Separation method provides a slightly improved result over Hierarchical Clustering usage. This is hypothesized to be the tuning of Hierarchical Clustering that favors flat clusters, hence not fitting the actual cluster shape of the data. Detailed numerical comparisons are included in Figure \ref{fig:ablation-table}. }
    \label{fig:cor-dim-reduction-results}
\end{figure}
Fig. \ref{fig:cor-dim-reduction-results} shows that the Cross-table Connecting method improves synthetic data fidelity over both the DEREC and direct flattening benchmarks \cite{derec}. Despite differing independence setups, Fig. \ref{fig:cor-dim-reduction-results} shows that the method consistently outperforms the benchmarks, demonstrating its reliability. Direct flattening creates a denser middle distribution compared to DEREC, indicating that merging child tables adds noisy information, which harms the data more than treating them independently. Even with a different similarity measure, the three setups show consistent outperformance. Both evaluation approaches suggest that cross-table data collaboration enhances the synthesizer's ability to capture the original data structure.

%Fig. \ref{fig:cor-dim-reduction-results} shows the Cross-table Connecting method bring improvement to synthetic data fidelity over both the DEREC and the direct flattening benchmark \cite{derec}. Despite the setups to determine independence are different, Fig. \ref{fig:cor-dim-reduction-results} shows that the distribution of the three setups consistently outperform the benchmarks, indicating consistency of this method. It is shown that directly flattening creates similarity distribution to have a denser area in the middle compared to the DEREC benchmark, indicating that the additional information created by direct flattening two child tables is noisy and brings more harm than benefit compared to treating them independent. Even if a different similarity approach, i.e. the W-distance, is used, the three setups also show traits of outperformance. The distribution is considered to be more similar if the distance is smaller, where all three distributions show a denser area at the beginning and a lighter tail after $x \geq 0.1$. With two different evaluation approaches giving similar insights, this can be concluded that data collaboration across tables indeed brings additional knowledge so that the synthesizer can capture the original data structure better.  
\subsubsection{Effects of different Cross-table Connecting setups}
Diving deeper to compare between results of the three Cross-table Connecting setups, Fig. \ref{fig:cor-dim-reduction-results} shows that segregating the independent columns using the 'up-and-stay' Threshold Separation method provides a slightly more consistent result compared to using Hierarchical Clustering, with a heavier tail at the right end. This may be due to the tuning of Hierarchical Clustering that favors creating flat clusters, which oversimplifies the clustering step. However, in terms of W-distance, it is shown that the Hierarchical Clustering method yields a slightly better fidelity by having a denser area close to 0. This performance difference is explained by the underlying setup of Hierarchical Clustering, as features are separated based on their pairwise distances, hence favoring distance-based evaluation. 

 \subsection{Ablation Study} 
 \begin{figure}[H]
    \centering
    \includegraphics[width = 1 \linewidth]{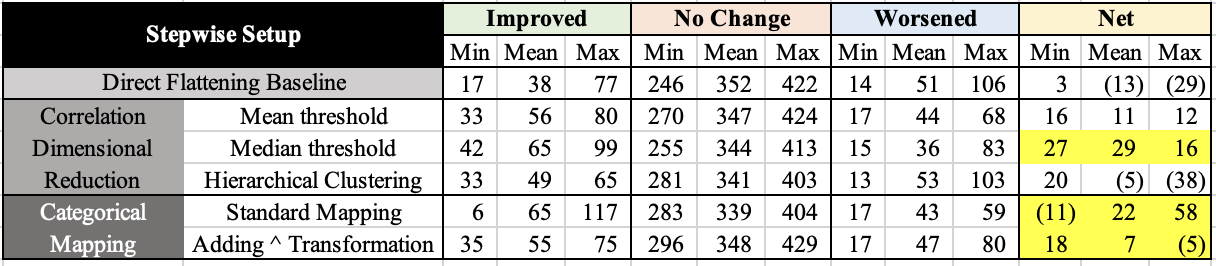}
    \caption{The list contains the max, min and average counts of performance across the eight individual trials. The Data Semantic Enhancement Method (Sec. \ref{sec: data-semantic-enhacement-system}) improves synthetic fidelity with a significant improvement on the average and max value (the higher end of the distribution). The Cross-table Connecting Method (Sec. \ref{sec: cross-table-connecting-method}) shows consistent net improvement with more improved and less worsened number in all three counts.}
    \label{fig:ablation-table}
\end{figure}
The GReaTER model consists of two main components: Data Semantic Enhancement System and Cross-table Connecting Method. The ablation study aims to measure the stepwise change in data fidelity when each method is applied. Using direct flattening as the baseline, Table Fig. \ref{fig:ablation-table} shows that Cross-table Connecting Method, when optimized, leads to a consistent net fidelity improvement across number of improved pairwise column fidelity. When fixing this method to assess the effect of the Data Semantic Enhancement System, net fidelity improvement is also observed.\\
\\
The standard Data Semantic Enhancement System (Sec. \ref{sec: data-semantic-enhacement-system}) significantly improves the mean and maximum fidelity of column pairs, while the special transformation (Sec. \ref{subsub: special-transform}) primarily benefits the lower end of the fidelity distribution. This difference is hypothesized to stem from semantic effects. The standard mapping helps establish clearer relationships between appropriately tokenized tokens. It is hypothesized that poorly tokenized tokens are disregarded by the LLM, as the model focuses on improving those columns that it can relate. In contrast, the data-specific transformation enables more values to be appropriately tokenized, allowing previously uninterpretable columns to become understandable, thus improving the lower end. However, this also risks over-establishing correlations between features, which may confuse the LLM and disrupt its understanding of originally interpretable feature correlations. Despite these differences, both modules in Data Semantic Enhancement System improve performance, highlighting the importance of natural-language-like data in facilitating LLM comprehension.

\section{Conclusion and Future Work}
\label{sec: further-discussion}
In this work, we study the unexplored problem of how to effectively leverage pre-trained knowledge of Large Language Model to complete quality multi-modal and multi-tabular data synthesis. We propose an effective model GReaTER, which outperforms the state-of-the-art approaches in the tabular data generation area. In GReaTER, we first perform parent table extraction for structural data and multi-tabular data. In addition, the Data Semantic Enhancement System is used to generate more semantically meaningful labels for categorical features. Next, tabular features are flattened using the Cross-table Connecting Method to restructure complex multi-table structure into a one-to-many relationship. Extensive experiments demonstrate the rationality and effectiveness of GReaTER.\\ 
\\
In future work, we would like to improve quality of in-context learning via utilizing more advanced LLMs. How to effectively fine-tune LLM for more precise tabular data generation, still awaits further exploration. Specifically, we see three implications for future work, (1) \textbf{Less effective when datasets are highly correlated:} The Cross-table Connecting Method (Sec. \ref{sec: cross-table-connecting-method}) may be less effective when dealing with data from sectors with highly correlated features (e.g. financial data), so that a more robust feature clustering method can be worked on, (2) \textbf{Extension of semantic enhancement in LLM applications}: The significance of input semantics reflected by the Data Semantics Enhancement System opens up further applications in LLM use cases, such as automating the understandability-based transformation module using LLM to generate more precise categories. Other than updating tabular categories with precision, a more robust textual encoding framework can also be automated, for instance, a system that textual-encodes row observations differently, from 'Name: Grace, Gender: Female, Lunch: Rice, Dinner: Steak, Access Device: Laptop, Genre: Action" in Fig. \ref{fig:GReaT_implementation_problem} to a more specific version like "A female named Grace had rice for lunch and steak for dinner while watching action-related video with laptop.", whose additional sentence semantic may further improve understanding of LLM, and (3) \textbf{Incorporation of more advanced LLMs}: Potential improvement is foreseen by incorporating more advanced LLMs as synthesizer backbone, given the synthetic performance shown in advanced LLMs such as GPT-4o \cite{openai2024gpt4technicalreport} and Llama 3.2 \cite{llama32}. This usage of out-dated LLM can also be reflected by its limited outperformance of understandability-based transformation module as the model does not have a pre-trained knowledge base equally diverse. With advanced LLM showing improvement in its ability to understand text, we expect the enhancement in text semantics can be further leveraged to process and generate tabular synthetic data of higher quality.

\section*{Acknowledgment}
This research is supported by NSF -- CNS (2247795), Office of Naval Research, (ONR N00014-22-1-2680) and CISCO Research Grant. Any opinions, findings, and conclusions expressed in this material are those of the authors and do not reflect the views of the National Science Foundation or the Office of Naval Research.

\bibliographystyle{plain}
%%% abbrv, alpha, acm, siam, apalike, plain
\nocite{*}
\bibliography{sample-base}

\clearpage

\appendix
\appendix
\section{Definition}
\subsection{Definition of Parent and Child Table} \label{def:parent-and-child}
A parent table $\textbf{T}$ contains different observations with $n$ features so for each observation $o$ from unique identifier $i$, $o_{i} = [x_{i1}, x_{i2}, \dots, x_{in}]$. A child table \textbf{s} is a table containing some observations on every unique parental observation $o_{i}$. A child table subset that contains observations of parent $o_{i}$ only can be expressed as: 
\begin{equation} \label{eq:parent_notation_simp}
    s_{o_{i}} = \begin{bmatrix}
    o^{(1)}_{i} \\ 
    o^{(2)}_{i} \\
    \cdots \\
    o^{(k)}_{i}
\end{bmatrix} = \begin{bmatrix}
    x'^{(1)}_{i1} & x'^{(1)}_{i2} & \cdots & x'^{(1)}_{im}\\
    x'^{(2)}_{i1} & x'^{(2)}_{i2} & \cdots & x'^{(2)}_{im}\\
    \cdots \\
    x'^{(k)}_{i1} & x'^{(k)}_{i2} & \cdots & x'^{(k)}_{im}
\end{bmatrix}.
\end{equation}
Therefore, the child table consisting of child observations from every parental subject from 1 to $l$ can be expressed as follow: 
\begin{equation} \label{eq:whole-child-notation}
s = \begin{bmatrix}
    o^{(1)}_{1} & o^{(2)}_{1} & o^{(3)}_{1} & \cdots & o^{(1)}_{2} & o^{(2)}_{2} & \cdots & o^{(1)}_{l} & \cdots
\end{bmatrix}'.
\end{equation}
%Given the $j$-th column of all observations on the child table subset under $o_{i}$ observations containing the same value, i.e. $x'^{(p)}_{ij} = x'^{(q)}_{ij}, 1 \leq p\neq q \leq k$, this $j$-th column is regarded as a contextual information for child table subset of $o_{i}$, i.e. $x'_{ij} \in C^{i}$.\\

\begin{figure}[H]
    \centering
    \includegraphics[width = 0.6 \linewidth]{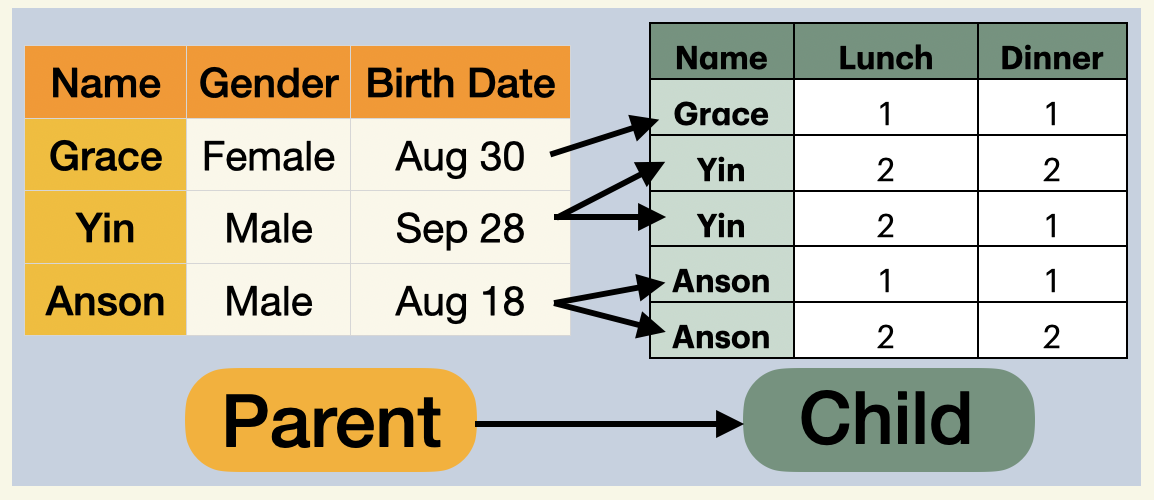}
    \caption{Multi-table synthesizers require a parent-child table hierarchy (one-to-many table relationship). The table containing \textbf{one} occurrence per unique subject will serve as the parent while the table containing \textbf{many} occurrences per unique subject will serve as child.}
    \label{fig:one-to-many-combine}
\end{figure}

\textit{Example}: A parent table contains one row to record attribute of each user, such as the yellow membership table in Fig. \ref{fig:one-to-many-combine}. In contrast, a child table contains multiple observations from each unique user, like the visit logbook recording every visit of each user in Fig. \ref{fig:one-to-many-combine}. 
\subsection{Definition of Contextual Variables} \label{def:contextual-variable-notations}
Contextual variables, which are columns that remain constant within a sequence \cite{sdv}, are utilized in this work to create a parent table following the DEREC pipeline \cite{derec}. Writing equation for entire child table (Eq. \ref{eq:whole-child-notation}) in a detailed manner as Eq. \ref{eq:parent_notation_simp}, the child table can be further expressed as follows: 

\begin{equation}\label{eq:parent_notation_detail}
    s = \begin{bmatrix}
    o^{(1)}_{1} \\
    o^{(2)}_{1} \\
    o^{(3)}_{1} \\
    \cdots \\
    o^{(1)}_{2} \\
    o^{(2)}_{2} \\
    o^{(3)}_{2} \\
    \cdots \\
    o^{(1)}_{l} \\
    \cdots
\end{bmatrix} = \begin{bmatrix}
    x'^{(1)}_{11} & x'^{(1)}_{12} & \cdots & x'^{(1)}_{1m} \\
    x'^{(2)}_{11} & x'^{(2)}_{12} & \cdots & x'^{(2)}_{1m} \\
    x'^{(3)}_{11} & x'^{(3)}_{12} & \cdots & x'^{(3)}_{1m} \\
    & \cdots \\
    x'^{(1)}_{21} & x'^{(1)}_{22} & \cdots & x'^{(1)}_{2m} \\
    x'^{(2)}_{21} & x'^{(2)}_{22} & \cdots & x'^{(2)}_{2m} \\
    x'^{(3)}_{21} & x'^{(3)}_{22} & \cdots & x'^{(3)}_{2m} \\
    & \cdots \\
    x'^{(1)}_{l1} & x'^{(1)}_{l2} & \cdots & x'^{(1)}_{lm} \\
    & \cdots 
\end{bmatrix}.
\end{equation}

Consider the column $x'_{n}$
\begin{align*}
    x'_{n}&=\begin{bmatrix}
    x'^{(1)}_{1n} &
    x'^{(2)}_{1n} &
    \cdots &
    x'^{(1)}_{2n} &
    x'^{(2)}_{2n} &
    \cdots &
    x'^{(1)}_{ln} &
    \cdots 
\end{bmatrix}'\\ 
&=\begin{bmatrix}
    \vec{x'_{1n}} &
    \vec{x'_{2n}} &
    \cdots &
    \vec{x'_{ln}} &
    \cdots
\end{bmatrix}'.
\end{align*} 

If the $n$-th column value for the unique identifier $i$ is consistent for every $i$-related observation, $x'^{(a)}_{in} = x'^{(b)}_{in} \forall a \neq b$, then $\vec{x'_{in}}$ is contextual for unique identifier $i$. If $m(<100\%$ due to realistic exceptional cases and measurement error) of unique subjects contain contextual information in the column, this column is considered contextual. A parent table containing contextual columns $=[1, 2, \cdots, g]$ is 
$$
T = \begin{bmatrix}
    x'_{11} & x'_{12} & & x'_{1g} \\
    x'_{21} & x'_{22} & \cdots &  x'_{2g}\\
    \cdots \\
    x'_{l1} & x'_{l2} & & x'_{lg}\\
\end{bmatrix}
$$ for $x'^{(a)}_{11} = x'^{(b)}_{11}$, $x'^{(a)}_{12} = x'^{(b)}_{12}$, $\cdots$, $x'^{(a)}_{1g} = x'^{(b)}_{1g}$ for $a \neq b$

\textit{Example}:
 Fig. \ref{fig:one-to-many-combine} shows that the gender and birth date of the same subject in different visits should be consistent across different visits, because these two features generally come with one's birth, and is rarely changed in a big-data perspective. 
\begin{figure}[H]
    \centering
    \includegraphics[width = 0.6 \linewidth]{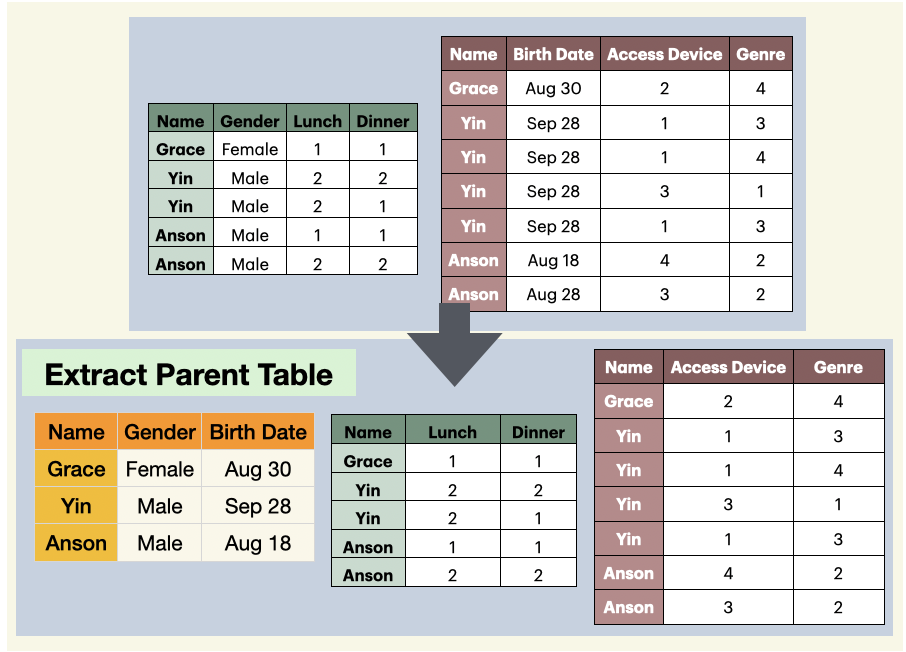}
    \caption{Despite having multiple visit records for each subject, the gender and birth date do not vary across visits, so that these contextual variables can be extracted to create a table form compliant to the parent and child table structure.}
    \label{fig:many-to-many-combine-with-derec}
\end{figure} 

\section{Cross-table Feature Correlation Algorithm}
\begin{algorithm}[H]
\caption{Cross-table Feature Correlation computation}\label{algo:conditional_distribution_indicator}
\begin{enumerate}
    \item Given original columns 1 and 2 $\vec{x}_{1;O}, \vec{x}_{2;O}=\begin{pmatrix}
    x^{1;O}_{1} & x^{2;O}_{2} & \cdots
\end{pmatrix}^{\top}, \begin{pmatrix}
    x^{2;O}_{1} & x^{2;O}_{2} & \cdots
\end{pmatrix}^{\top}$, and synthetic columns 1 and 2 $\vec{x}_{1;syn}, \vec{x}_{2;syn} =\begin{pmatrix}
    x^{1;syn}_{1} & x^{1;syn}_{2} & \cdots
\end{pmatrix}^{\top}$, $ \begin{pmatrix}
    x^{2;syn}_{1} & x^{2;syn}_{2} & \cdots
\end{pmatrix}^{\top}$
    \item Measure $P(x^{2}_{j}|x^{1}_{i})$ of every $x^{2}_{j}$ to obtain the conditional distribution of $x^{2}|x^{1}_{i}$
    \item Implement (2) on both the original and synthetic data to obtain $x^{2;O}|x^{1;O}_{i}$ and $x^{2;syn}|x^{1;syn}_{i}$
    \item Measure distribution similarity with to quantify a similarity indicator given column 1 value $x^{1}_{i}$ 
    \item Repeat (2) to (4) on every column 1 value $x^{1}_{i}$ to obtain a conditional distribution given every column 1 value
    \item Compute cross-table feature correlation $Z_{x^{2}|x^{1}}$ by taking the weighted average of all $Z_{x^{2}|x^{1}_{i}}$ with the probability of the occurrence for every unique parent value $P(x^{1}_{i})$
    \item Repeat on all column pair for the similarity distribution
\end{enumerate}
\end{algorithm}

\end{document}

%% file: def-smile.tex
\numberwithin{example}{section}

%% file: front.tex
\usepackage[utf8x]{inputenc}
\usepackage{mathrsfs}
\usepackage{amssymb}
\usepackage{amsfonts}
\usepackage{amsmath}
\usepackage{amsthm,verbatim}
\usepackage{wrapfig}
\usepackage{multirow}
\usepackage{longtable}
\usepackage{enumerate}
\usepackage{color, graphicx}
\usepackage{tikz}

\def\boxit#1{\vbox{\hrule\hbox{\vrule\kern6pt
          \vbox{\kern6pt#1\kern6pt}\kern6pt\vrule}\hrule}}

%% file: main.bbl
\begin{thebibliography}{10}

\bibitem{privacy}
S.~A. Abdelhameed, S.~M. Moussa, and M.~E. Khalifa.
\newblock Privacy-preserving tabular data publishing: a comprehensive evaluatoin from web to cloud.
\newblock {\em Computers \& Security}, 72, 2018.

\bibitem{Abril07}
Patricia~S. Abril and Robert Plant.
\newblock The patent holder's dilemma: Buy, sell, or troll?
\newblock {\em Communications of the ACM}, 50(1):36--44, January 2007.

\bibitem{alimohammadi2024adapting}
Kaveh Alimohammadi, Hao Wang, Ojas Gulati, Akash Srivastava, and Navid Azizan.
\newblock Adapting differentially private synthetic data to relational databases.
\newblock {\em arXiv preprint: 2405.18670}, 2024.

\bibitem{anonymous2024tabmeta}
Anonymous.
\newblock Tabmeta: Table metadata generation with {LLM}-curated dataset and {LLM}-judges.
\newblock In {\em Submitted to ACL Rolling Review - June 2024}, 2024.
\newblock under review.

\bibitem{claude3opus}
Anthropic.
\newblock Introducing the next generation of claude.
\newblock \url{https://www.anthropic.com/news/claude-3-family}, 2024.
\newblock Accessed: 2024-03-27.

\bibitem{Baldin_Cannon_2024}
Nicolai Baldin and Tom Cannon.
\newblock Synthesized uses gen ai for compliant bigquery dataset snapshots | google cloud blog.
\newblock {\em Google Data Analytics}, Feb 2024.

\bibitem{bayat2023a}
Reza Bayat.
\newblock A study on sample diversity in generative models: {GAN}s vs. diffusion models.
\newblock {\em ICLR 2023 Tiny Papers}, 2023.

\bibitem{llm}
Yoshua Bengio, Réjean Ducharme, Pascal Vincent, and Christian Jauvin.
\newblock A neural probabilistic language model.
\newblock {\em Journal of machine learning research}, 3, 2002.

\bibitem{benjamini-hochberg}
Yoav Benjamini and Yosef Hochberg.
\newblock Controlling the false discovery rate: A practical and powerful approach to multiple testing.
\newblock {\em Journal of the Royal Statistical Society. Series B (Methodological)}, 57(1):289--300, 1995.

\bibitem{borisov2023language}
Vadim Borisov, Kathrin Sessler, Tobias Leemann, Martin Pawelczyk, and Gjergji Kasneci.
\newblock Language models are realistic tabular data generators.
\newblock In {\em The Eleventh International Conference on Learning Representations}, 2023.

\bibitem{gpt3}
Tom Brown, Benjamin Mann, Nick Ryder, Melanie Subbiah, Jared~D Kaplan, Prafulla Dhariwal, Arvind Neelakantan, Pranav Shyam, Girish Sastry, Amanda Askell, Sandhini Agarwal, Ariel Herbert-Voss, Gretchen Krueger, Tom Henighan, Rewon Child, Aditya Ramesh, Daniel Ziegler, Jeffrey Wu, Clemens Winter, Chris Hesse, Mark Chen, Eric Sigler, Mateusz Litwin, Scott Gray, Benjamin Chess, Jack Clark, Christopher Berner, Sam McCandlish, Alec Radford, Ilya Sutskever, and Dario Amodei.
\newblock Language models are few-shot learners.
\newblock In H.~Larochelle, M.~Ranzato, R.~Hadsell, M.F. Balcan, and H.~Lin, editors, {\em Advances in Neural Information Processing Systems}, volume~33, pages 1877--1901. Curran Associates, Inc., 2020.

\bibitem{extract_from_diffusion}
Nicholas Carlini, Jamie Hayes, Milad Nasr, Matthew Jagielski, Vikash Sehwag, Florian Tram\`{e}r, Borja Balle, Daphne Ippolito, and Eric Wallace.
\newblock Extracting training data from diffusion models.
\newblock In {\em Proceedings of the 32nd USENIX Conference on Security Symposium}, SEC '23, USA, 2023. USENIX Association.

\bibitem{smote}
Nitesh~V Chawla, Kevin~W Bowyer, Lawrence~O Hall, and W~Philip Kegelmeyer.
\newblock Smote: synthetic minority over-sampling technique.
\newblock {\em Journal of artificial intelligence research}, 16, 2002.

\bibitem{attack}
L.~Cheng, F.~Liu, and D.~Yao.
\newblock Enterprise data breach: causes, challenges, prevention, and future directions.
\newblock {\em Wiley Interdiscilpinary Reviews: Data Mining and Knowledge Discovery}, 7(5), 2017.

\bibitem{cheng2024downstream}
Yinan Cheng, Chi-Hua Wang, Vamsi~K Potluru, Tucker Balch, and Guang Cheng.
\newblock Downstream task-oriented generative model selections on synthetic data training for fraud detection models.
\newblock {\em arXiv preprint arXiv:2401.00974}, 2024.

\bibitem{MedGAN}
Edward Choi, Siddharth Biswal, Bradley Malin, Jon Duke, Walter~F. Stewart, and Jimeng Sun.
\newblock Generating multi-label discrete patient records using generative adversarial networks.
\newblock {\em Machine Learning for Healthcare Conference (PMLR)}, 2017.

\bibitem{Cohen07}
Sarah Cohen, Werner Nutt, and Yehoshua Sagic.
\newblock Deciding equivalances among conjunctive aggregate queries.
\newblock {\em J. ACM}, 54(2), April 2007.

\bibitem{cohenaddad2017hierarchicalclusteringobjectivefunctions}
Vincent Cohen-addad, Varun Kanade, Frederik Mallmann-trenn, and Claire Mathieu.
\newblock Hierarchical clustering: Objective functions and algorithms.
\newblock {\em J. ACM}, 66(4), June 2019.

\bibitem{sdmetrics}
DataCebo, Inc.
\newblock {\em Synthetic Data Metrics}, 10 2023.
\newblock Version 0.12.0.

\bibitem{numerical-linear-algebra}
Biswa~Nath Datta.
\newblock {\em Numerical Linear Algebra and Applications, Second Edition}.
\newblock Society for Industrial and Applied Mathematics, USA, 2nd edition, 2010.

\bibitem{Desai2021OnRO}
Dhruv Desai and Dhagash Mehta.
\newblock On robustness of mutual funds categorization and distance metric learning.
\newblock In {\em The Journal of Financial Data Science}, 2021.

\bibitem{dhariwal2021diffusion}
Prafulla Dhariwal and Alexander~Quinn Nichol.
\newblock Diffusion models beat {GAN}s on image synthesis.
\newblock In A.~Beygelzimer, Y.~Dauphin, P.~Liang, and J.~Wortman Vaughan, editors, {\em Advances in Neural Information Processing Systems}, 2021.

\bibitem{count1}
Wayne Fife.
\newblock {\em The Importance of Counting for Qualitative Research}, pages 121--128.
\newblock Springer International Publishing, Cham, 2020.

\bibitem{kstest}
Jr. Frank J.~Massey.
\newblock The kolmogorov-smirnov test for goodness of fit.
\newblock {\em Journal of the American Statistical Association}, 46(253):68--78, 1951.

\bibitem{GARCIA2015108}
Luís~P.F. Garcia, André~C.P.L.F. {de Carvalho}, and Ana~C. Lorena.
\newblock Effect of label noise in the complexity of classification problems.
\newblock {\em Neurocomputing}, 160:108--119, 2015.

\bibitem{disease}
Mauro Giuffrè and Dennis~L. Shung.
\newblock Harnessing the power of synthetic data in healthcare: innovation, application, and privacy.
\newblock {\em Digital Medicine}, 2023.

\bibitem{Hagerup1993}
Torben Hagerup, Kurt Mehlhorn, and J.~Ian Munro.
\newblock Maintaining discrete probability distributions optimally.
\newblock In {\em Proceedings of the 20th International Colloquium on Automata, Languages and Programming}, volume 700 of {\em Lecture Notes in Computer Science}, pages 253--264, Berlin, 1993. Springer-Verlag.

\bibitem{count2}
David Hannah and Brenda Lautsch.
\newblock Counting in qualitative research: Why to conduct it, when to avoid it, and when to closet it.
\newblock {\em Journal of Management Inquiry - J MANAGE INQUIRY}, 20:14--22, 03 2011.

\bibitem{herbrich2022data}
Tilman Herbrich.
\newblock Data clean rooms.
\newblock {\em Computer Law Review International}, 23(4):109--120, 2022.

\bibitem{hsieh2024improve}
Din-Yin Hsieh, Chi-Hua Wang, and Guang Cheng.
\newblock Improve fidelity and utility of synthetic credit card transaction time series from data-centric perspective.
\newblock {\em arXiv preprint arXiv:2401.00965}, 2024.

\bibitem{utilizing-imperfect-synthetic}
Ting-Yao Hu, Mohammadreza Armandpour, Ashish Shrivastava, Jen-Hao~Rick Chang, Hema Koppula, and Oncel Tuzel.
\newblock Utilizing imperfect synthetic data to improve speech recognition.
\newblock In {\em ICASSP}, 2022.

\bibitem{hunner2016names}
Trey Hunner and Simeon Visser.
\newblock names: A python library for generating random names.
\newblock \url{https://github.com/treyhunner/names}, 2016.
\newblock Accessed: 2024-03-27.

\bibitem{rel_vs_non_rel}
N.~Jatana, S.~Puri, M.~Ahuja, I.~Kathuria, and D.~Gosain.
\newblock A survey and comparison of relational and non-relational database.
\newblock {\em International Journal of Engineering Research \& Technology}, 2012.

\bibitem{ji2014differential}
Zhanglong Ji, Zachary~Chase Lipton, and Charles~Peter Elkan.
\newblock Differential privacy and machine learning: a survey and review.
\newblock {\em ArXiv}, abs/1412.7584, 2014.

\bibitem{modular}
Ying Jin and Dominik Rothenh\"{a}usler.
\newblock Modular regression: improving linear models by incorporating auxiliary data.
\newblock {\em J. Mach. Learn. Res.}, 24(1), January 2023.

\bibitem{Jolliffe:1986}
I.~T. Jolliffe and J.~Cadima.
\newblock Principal component analysis: A review and recent developments.
\newblock {\em Philosophical Transactions. Series A, Mathematical, Physical, and Engineering Sciences}, 374(2065):20150202, 2016.

\bibitem{inbook}
Michael~W. Kearney.
\newblock Cramér's v.
\newblock In M.~R. Allen, editor, {\em Sage Encyclopedia of Communication Research Methods}, page n107. Sage, 2017.

\bibitem{traffic}
Dipika Khullar, Yash Shah, Ninad Kulkarni, and Negin Sokhandan.
\newblock Synthetic data generation for scarce road scene detection scenarios.
\newblock In {\em NeurIPS Workshop on Synthetic Data Generation with Generative AI}, 2023.

\bibitem{ref1}
Wilhelm Kirch, editor.
\newblock {\em Pearson's Correlation Coefficient}, pages 1090--1091.
\newblock Springer Netherlands, Dordrecht, 2008.

\bibitem{wdis2}
Soheil Kolouri.
\newblock Optimal mass transport: Signal processing and machine-learning applications.
\newblock {\em IEEE Signal Processing Magazine}, 34(4):43 -- 59, 2017.

\bibitem{tabddpm}
Akim Kotelnikov, Dmitry Baranchuk, Ivan Rubachev, and Artem Babenko.
\newblock Tabddpm: modelling tabular data with diffusion models.
\newblock In {\em Proceedings of the 40th International Conference on Machine Learning}, ICML'23. JMLR.org, 2023.

\bibitem{krishna2023posthocexplanationslanguage}
Satyapriya Krishna, Jiaqi Ma, Dylan Slack, Asma Ghandeharioun, Sameer Singh, and Himabindu Lakkaraju.
\newblock Post hoc explanations of language models can improve language models.
\newblock In {\em Proceedings of the 37th International Conference on Neural Information Processing Systems}, NIPS '23, Red Hook, NY, USA, 2023. Curran Associates Inc.

\bibitem{derec}
Tung Sum~Thomas Kwok, Chi-Hua Wang, and Guang Cheng.
\newblock {DEREC-SIMPRO: Unlock Language Model Benefits to Advance Synthesis in Data Clean Room}.
\newblock {\em ACM ICAIF Workshop arXiv:2411.00879}, 2024.

\bibitem{wdis}
Junior Leo, Ernest Ge, and Stotle Li.
\newblock Wasserstein distance in deep learning.
\newblock {\em SSRN Electron. J.}, 2023.

\bibitem{li2021online}
Yuantong Li, Chi-Hua Wang, and Guang Cheng.
\newblock Online forgetting process for linear regression models.
\newblock In {\em International Conference on Artificial Intelligence and Statistics}, pages 217--225. PMLR, 2021.

\bibitem{liu2022utility}
Yucong Liu, Chi-Hua Wang, and Guang Cheng.
\newblock On the utility recovery incapability of neural net-based differential private tabular training data synthesizer under privacy deregulation.
\newblock {\em arXiv preprint arXiv:2211.15809}, 2022.

\bibitem{detect_data_copying}
C.~Meehan, K.~Chaudhuri, and S.~Dasgupta.
\newblock A nonparametric test to detect data-copying in generative models.
\newblock {\em International Conference on Artificial Intelligence and Statistics}, 2020.

\bibitem{meehan2020nonparametric}
Casey Meehan, Kamalika Chaudhuri, and Sanjoy Dasgupta.
\newblock A non-parametric test to detect data-copying in generative models.
\newblock {\em ArXiv}, abs/2004.05675, 2020.

\bibitem{mendelevitch2021fidelity}
Ofer Mendelevitch and Michael~D. Lesh.
\newblock Fidelity and privacy of synthetic medical data.
\newblock {\em ArXiv}, abs/2101.08658, 2021.

\bibitem{llama32}
Meta.
\newblock Introducing llama 3.2.
\newblock \url{https://www.llama.com/}, 2024.
\newblock Accessed: 2024-03-27.

\bibitem{müllner2011modern}
Daniel M{\"u}llner.
\newblock Modern hierarchical, agglomerative clustering algorithms.
\newblock {\em ArXiv}, abs/1109.2378, 2011.

\bibitem{Mumuni_2024}
Alhassan Mumuni, Fuseini Mumuni, and Nana~Kobina Gerrar.
\newblock A survey of synthetic data augmentation methods in machine vision.
\newblock {\em Machine Intelligence Research}, 21(5):831–869, March 2024.

\bibitem{confident_learning}
Curtis~G. Northcutt, Lu~Jiang, and Isaac~L. Chuang.
\newblock Confident learning: Estimating uncertainty in dataset labels.
\newblock {\em Journal of Artificial Intelligence Research (JAIR)}, 2021.

\bibitem{NVIDIA}
NVIDIA.
\newblock Use case: Synthetic data generation.
\newblock \url{https://www.nvidia.com/en-us/use-cases/synthetic-data/}, n.d.
\newblock Accessed: 2024-03-27.

\bibitem{gpt35}
OpenAI.
\newblock Gpt-3.5 turbo.
\newblock \url{https://platform.openai.com/docs/models/gpt-3-5-turbo}, 2024.
\newblock Accessed: 2024-03-27.

\bibitem{gpt4o}
OpenAI.
\newblock Introducing gpt-4o and more tools to chatgpt free users.
\newblock {\em OpenAI Blog}, 2024.

\bibitem{openai2024gpt4technicalreport}
OpenAI, Josh Achiam, Steven Adler, Sandhini Agarwal, Lama Ahmad, Ilge Akkaya, Florencia~Leoni Aleman, Diogo Almeida, Janko Altenschmidt, Sam Altman, Shyamal Anadkat, Red Avila, Igor Babuschkin, Suchir Balaji, Valerie Balcom, Paul Baltescu, Haiming Bao, Mohammad Bavarian, Jeff Belgum, Irwan Bello, Jake Berdine, Gabriel Bernadett-Shapiro, Christopher Berner, Lenny Bogdonoff, Oleg Boiko, Madelaine Boyd, Anna-Luisa Brakman, Greg Brockman, Tim Brooks, Miles Brundage, Kevin Button, Trevor Cai, Rosie Campbell, Andrew Cann, Brittany Carey, Chelsea Carlson, Rory Carmichael, Brooke Chan, Che Chang, Fotis Chantzis, Derek Chen, Sully Chen, Ruby Chen, Jason Chen, Mark Chen, Ben Chess, Chester Cho, Casey Chu, Hyung~Won Chung, Dave Cummings, Jeremiah Currier, Yunxing Dai, Cory Decareaux, Thomas Degry, Noah Deutsch, Damien Deville, Arka Dhar, David Dohan, Steve Dowling, Sheila Dunning, Adrien Ecoffet, Atty Eleti, Tyna Eloundou, David Farhi, Liam Fedus, Niko Felix, Simón~Posada Fishman, Juston Forte, Isabella Fulford, Leo
  Gao, Elie Georges, Christian Gibson, Vik Goel, Tarun Gogineni, Gabriel Goh, Rapha Gontijo-Lopes, Jonathan Gordon, Morgan Grafstein, Scott Gray, Ryan Greene, Joshua Gross, Shixiang~Shane Gu, Yufei Guo, Chris Hallacy, Jesse Han, Jeff Harris, Yuchen He, Mike Heaton, Johannes Heidecke, Chris Hesse, Alan Hickey, Wade Hickey, Peter Hoeschele, Brandon Houghton, Kenny Hsu, Shengli Hu, Xin Hu, Joost Huizinga, Shantanu Jain, Shawn Jain, Joanne Jang, Angela Jiang, Roger Jiang, Haozhun Jin, Denny Jin, Shino Jomoto, Billie Jonn, Heewoo Jun, Tomer Kaftan, Łukasz Kaiser, Ali Kamali, Ingmar Kanitscheider, Nitish~Shirish Keskar, Tabarak Khan, Logan Kilpatrick, Jong~Wook Kim, Christina Kim, Yongjik Kim, Jan~Hendrik Kirchner, Jamie Kiros, Matt Knight, Daniel Kokotajlo, Łukasz Kondraciuk, Andrew Kondrich, Aris Konstantinidis, Kyle Kosic, Gretchen Krueger, Vishal Kuo, Michael Lampe, Ikai Lan, Teddy Lee, Jan Leike, Jade Leung, Daniel Levy, Chak~Ming Li, Rachel Lim, Molly Lin, Stephanie Lin, Mateusz Litwin, Theresa Lopez, Ryan
  Lowe, Patricia Lue, Anna Makanju, Kim Malfacini, Sam Manning, Todor Markov, Yaniv Markovski, Bianca Martin, Katie Mayer, Andrew Mayne, Bob McGrew, Scott~Mayer McKinney, Christine McLeavey, Paul McMillan, Jake McNeil, David Medina, Aalok Mehta, Jacob Menick, Luke Metz, Andrey Mishchenko, Pamela Mishkin, Vinnie Monaco, Evan Morikawa, Daniel Mossing, Tong Mu, Mira Murati, Oleg Murk, David Mély, Ashvin Nair, Reiichiro Nakano, Rajeev Nayak, Arvind Neelakantan, Richard Ngo, Hyeonwoo Noh, Long Ouyang, Cullen O'Keefe, Jakub Pachocki, Alex Paino, Joe Palermo, Ashley Pantuliano, Giambattista Parascandolo, Joel Parish, Emy Parparita, Alex Passos, Mikhail Pavlov, Andrew Peng, Adam Perelman, Filipe de~Avila Belbute~Peres, Michael Petrov, Henrique~Ponde de~Oliveira~Pinto, Michael, Pokorny, Michelle Pokrass, Vitchyr~H. Pong, Tolly Powell, Alethea Power, Boris Power, Elizabeth Proehl, Raul Puri, Alec Radford, Jack Rae, Aditya Ramesh, Cameron Raymond, Francis Real, Kendra Rimbach, Carl Ross, Bob Rotsted, Henri Roussez,
  Nick Ryder, Mario Saltarelli, Ted Sanders, Shibani Santurkar, Girish Sastry, Heather Schmidt, David Schnurr, John Schulman, Daniel Selsam, Kyla Sheppard, Toki Sherbakov, Jessica Shieh, Sarah Shoker, Pranav Shyam, Szymon Sidor, Eric Sigler, Maddie Simens, Jordan Sitkin, Katarina Slama, Ian Sohl, Benjamin Sokolowsky, Yang Song, Natalie Staudacher, Felipe~Petroski Such, Natalie Summers, Ilya Sutskever, Jie Tang, Nikolas Tezak, Madeleine~B. Thompson, Phil Tillet, Amin Tootoonchian, Elizabeth Tseng, Preston Tuggle, Nick Turley, Jerry Tworek, Juan Felipe~Cerón Uribe, Andrea Vallone, Arun Vijayvergiya, Chelsea Voss, Carroll Wainwright, Justin~Jay Wang, Alvin Wang, Ben Wang, Jonathan Ward, Jason Wei, CJ~Weinmann, Akila Welihinda, Peter Welinder, Jiayi Weng, Lilian Weng, Matt Wiethoff, Dave Willner, Clemens Winter, Samuel Wolrich, Hannah Wong, Lauren Workman, Sherwin Wu, Jeff Wu, Michael Wu, Kai Xiao, Tao Xu, Sarah Yoo, Kevin Yu, Qiming Yuan, Wojciech Zaremba, Rowan Zellers, Chong Zhang, Marvin Zhang, Shengjia
  Zhao, Tianhao Zheng, Juntang Zhuang, William Zhuk, and Barret Zoph.
\newblock Gpt-4 technical report.
\newblock \url{https://arxiv.org/abs/2303.08774}, 2024.
\newblock Accessed: 2024-03-27.

\bibitem{tabgpt}
Inkit Padhi, Yair Schiff, Igor Melnyk, Mattia Rigotti, Youssef Mroueh, Pierre Dognin, Jerret Ross, Ravi Nair, and Erik Altman.
\newblock Tabular transformers for modeling multivariate time series.
\newblock In {\em ICASSP 2021 - 2021 IEEE International Conference on Acoustics, Speech and Signal Processing (ICASSP)}, pages 3565--3569, 2021.

\bibitem{wdis3}
Victor~M. Panaretos and Yoav Zemel.
\newblock Statistical aspects of wasserstein distances.
\newblock {\em Annual Review of Statistics and Its Application}, 6(Volume 6, 2019):405--431, 2019.

\bibitem{TableGAN}
Noseong Park, Mahmoud Mohammadi, Kshitij Gorde, Sushil Jajodia, Hongkyu Park, and Youngmin Kim.
\newblock Data synthesis based on generative adversarial networks.
\newblock {\em International Conference on Very Large Data Bases}, 2018.

\bibitem{ParraMoyano2024DataCollaboration}
José Parra-Moyano, Karl Schmedders, and Alex~"Sandy" Pentland.
\newblock How data collaboration platforms can help companies build better ai.
\newblock {\em Harvard Business Review}, January 2024.
\newblock Updated February 01, 2024.

\bibitem{hbs-data-collaboration}
José Parra-Moyano, Karl Schmedders, and Alex~“Sandy” Pentland.
\newblock How data collaboration platforms can help companies build better ai.
\newblock {\em Harvard Business Review}, 2024.

\bibitem{sdv}
Neha Patki, Roy Wedge, and Kalyan Veeramachaneni.
\newblock The synthetic data vault.
\newblock In {\em 2016 IEEE International Conference on Data Science and Advanced Analytics (DSAA)}, pages 399--410, 2016.

\bibitem{wdis4}
Gopi Prasad.
\newblock Fixed point theorems via w-distance in relational metric spaces with an application.
\newblock {\em Filomat}, 34:1889--1898, 12 2020.

\bibitem{radford2019language}
Alec Radford, Jeff Wu, Rewon Child, David Luan, Dario Amodei, and Ilya Sutskever.
\newblock Language models are unsupervised multitask learners.
\newblock 2019.
\newblock Accessed: 2024-03-27.

\bibitem{ramdas2015wasserstein}
Aaditya Ramdas, Nicolás~García Trillos, and Marco Cuturi.
\newblock On wasserstein two-sample testing and related families of nonparametric tests.
\newblock {\em Entropy}, 19(2), 2017.

\bibitem{chisq}
Rakesh Rana.
\newblock Chi-square test and its application in hypothesis testing.
\newblock {\em Journal of the Practice of Cardiovascular Sciences}, 1(1), 2015.

\bibitem{bank}
Gonçalo Ribeiro.
\newblock Synthetic data applications in finance.
\newblock \url{https://www.forbes.com/councils/forbestechcouncil/2024/04/03/synthetic-data-applications-in-finance/}, 2024.
\newblock Accessed: 2024-03-27.

\bibitem{9343211}
Sara Salim, Nour Moustafa, and Benjamin Turnbull.
\newblock Privacy-encoding models for preserving utility of machine learning algorithms in social media.
\newblock In {\em 2020 IEEE 19th International Conference on Trust, Security and Privacy in Computing and Communications (TrustCom)}, pages 856--863, 2020.

\bibitem{onehotencoding}
Jamell Samuels.
\newblock One-hot encoding and two-hot encoding: An introduction.
\newblock \url{10.13140/RG.2.2.21459.76327}, 01 2024.
\newblock Accessed: 2024-03-27.

\bibitem{satone2021fund2vec}
Vipul Satone, Dhruv Desai, and Dhagash Mehta.
\newblock Fund2vec: mutual funds similarity using graph learning.
\newblock In {\em Proceedings of the Second ACM International Conference on AI in Finance}, ICAIF '21, New York, NY, USA, 2022. Association for Computing Machinery.

\bibitem{schmidt2024tokenizationcompression}
Craig~W Schmidt, Varshini Reddy, Haoran Zhang, Alec Alameddine, Omri Uzan, Yuval Pinter, and Chris Tanner.
\newblock Tokenization is more than compression.
\newblock In Yaser Al-Onaizan, Mohit Bansal, and Yun-Nung Chen, editors, {\em Proceedings of the 2024 Conference on Empirical Methods in Natural Language Processing}, pages 678--702, Miami, Florida, USA, November 2024. Association for Computational Linguistics.

\bibitem{shankar2024silofuse}
Aditya Shankar, Hans Brouwer, Rihan Hai, and Lydia Chen.
\newblock Silofuse: Cross-silo synthetic data generation with latent tabular diffusion models.
\newblock In {\em 2024 IEEE 40th International Conference on Data Engineering (ICDE)}, pages 110--123, 2024.

\bibitem{singh2024rethinkinginterpretabilityeralarge}
Chandan Singh, Jeevana~Priya Inala, Michel Galley, Rich Caruana, and Jianfeng Gao.
\newblock Rethinking interpretability in the era of large language models.
\newblock {\em ArXiv}, abs/2402.01761, 2024.

\bibitem{solatorio2023realtabformer}
Aivin~V. Solatorio and Olivier Dupriez.
\newblock Realtabformer: Generating realistic relational and tabular data using transformers.
\newblock {\em arXiv preprint arXiv:2302.02041}, 2023.

\bibitem{VeeGAN}
Akash Srivastava, Lazar Valkov, Chris Russell, Michael~U Gutmann, and Charles Sutton.
\newblock Veegan: Reducing mode collapse in gans using implicit variational learning.
\newblock {\em Advances in Neural Information Processing Systems}, 2017.

\bibitem{tao2024discriminative}
Lan Tao, Shirong Xu, Chi-Hua Wang, Namjoon Suh, and Guang Cheng.
\newblock Discriminative estimation of total variation distance: A fidelity auditor for generative data.
\newblock {\em arXiv preprint arXiv:2405.15337}, 2024.

\bibitem{touvron2023llama}
Hugo Touvron, Thibaut Lavril, Gautier Izacard, Xavier Martinet, Marie-Anne Lachaux, Timoth{\'e}e Lacroix, Baptiste Rozi{\`e}re, Naman Goyal, Eric Hambro, Faisal Azhar, Aur{\'e}lien Rodriguez, Armand Joulin, Edouard Grave, and Guillaume Lample.
\newblock Llama: Open and efficient foundation language models.
\newblock {\em ArXiv}, abs/2302.13971, 2023.

\bibitem{touvron2023llama2openfoundation}
Hugo Touvron, Louis Martin, Kevin~R. Stone, Peter Albert, Amjad Almahairi, Yasmine Babaei, Nikolay Bashlykov, Soumya Batra, Prajjwal Bhargava, Shruti Bhosale, Daniel~M. Bikel, Lukas Blecher, Cristian~Cant{\'o}n Ferrer, Moya Chen, Guillem Cucurull, David Esiobu, Jude Fernandes, Jeremy Fu, Wenyin Fu, Brian Fuller, Cynthia Gao, Vedanuj Goswami, Naman Goyal, Anthony~S. Hartshorn, Saghar Hosseini, Rui Hou, Hakan Inan, Marcin Kardas, Viktor Kerkez, Madian Khabsa, Isabel~M. Kloumann, A.~V. Korenev, Punit~Singh Koura, Marie-Anne Lachaux, Thibaut Lavril, Jenya Lee, Diana Liskovich, Yinghai Lu, Yuning Mao, Xavier Martinet, Todor Mihaylov, Pushkar Mishra, Igor Molybog, Yixin Nie, Andrew Poulton, Jeremy Reizenstein, Rashi Rungta, Kalyan Saladi, Alan Schelten, Ruan Silva, Eric~Michael Smith, R.~Subramanian, Xia Tan, Binh Tang, Ross Taylor, Adina Williams, Jian~Xiang Kuan, Puxin Xu, Zhengxu Yan, Iliyan Zarov, Yuchen Zhang, Angela Fan, Melissa Hall~Melanie Kambadur, Sharan Narang, Aur{\'e}lien Rodriguez, Robert Stojnic,
  Sergey Edunov, and Thomas Scialom.
\newblock Llama 2: Open foundation and fine-tuned chat models.
\newblock {\em ArXiv}, abs/2307.09288, 2023.

\bibitem{wang2024badgd}
Chi-Hua Wang and Guang Cheng.
\newblock Badgd: A unified data-centric framework to identify gradient descent vulnerabilities.
\newblock {\em arXiv preprint arXiv:2405.15979}, 2024.

\bibitem{wangfederated}
Chi-Hua Wang, Wenjie Li, and Guang Lin.
\newblock Federated high-dimensional online decision making.
\newblock {\em Transactions on Machine Learning Research}.

\bibitem{ward2024data}
Joshua Ward, Chi-Hua Wang, and Guang Cheng.
\newblock Data plagiarism index: Characterizing the privacy risk of data-copying in tabular generative models.
\newblock {\em arXiv preprint arXiv:2406.13012}, 2024.

\bibitem{nepal}
WHO.
\newblock Collaborating for an improved civil registration system to advance health and population data system in nepal.
\newblock {\em WHO Newsroom}, 2024.

\bibitem{xia2024advancing}
Yu~Xia, Chi-Hua Wang, Joshua Mabry, and Guang Cheng.
\newblock Advancing retail data science: Comprehensive evaluation of synthetic data.
\newblock {\em arXiv preprint arXiv:2406.13130}, 2024.

\bibitem{xia2024data}
Zhangjie Xia, Chi-Hua Wang, and Guang Cheng.
\newblock Data deletion for linear regression with noisy sgd.
\newblock {\em arXiv preprint arXiv:2410.09311}, 2024.

\bibitem{digix_global_ai_challenge_2022}
Xiaojiu1414.
\newblock Ctr prediction - 2022 digix global ai challenge.
\newblock \url{https://www.kaggle.com/datasets/xiaojiu1414/digix-global-ai-challenge}, 2022.
\newblock Accessed: 2024-03-27.

\bibitem{ctgan}
Lei Xu, Maria Skoularidou, Alfredo Cuesta-Infante, and Kalyan Veeramachaneni.
\newblock Modeling tabular data using conditional gan.
\newblock In {\em Advances in Neural Information Processing Systems}, 2019.

\bibitem{tgan}
Lei Xu and Kalyan Veeramachaneni.
\newblock Synthesizing tabular data using generative adversarial networks.
\newblock {\em ArXiv}, abs/1811.11264, 2018.

\bibitem{xu2024utilitytheorysyntheticdata}
Shirong Xu, Will~Wei Sun, and Guang Cheng.
\newblock Utility theory of synthetic data generation.
\newblock {\em arXiv preprint: 2305.10015}, 2024.

\bibitem{xu2024bridginggapdifferentvocabularies}
Yangyifan Xu, Jinliang Lu, and Jiajun Zhang.
\newblock Bridging the gap between different vocabularies for llm ensemble.
\newblock {\em arXiv preprint:2404.09492}, 2024.

\bibitem{p-val-sim1}
L.~N. Yaddanapudi.
\newblock The american statistical association statement on p-values explained.
\newblock {\em Journal of Anaesthesiology, Clinical Pharmacology}, 32(4):421--423, 2016.

\bibitem{ye2023comprehensive}
Junjie Ye, Xuanting Chen, Nuo Xu, Can Zu, Zekai Shao, Shichun Liu, Yuhan Cui, Zeyang Zhou, Chao Gong, Yang Shen, Jie Zhou, Siming Chen, Tao Gui, Qi~Zhang, and Xuanjing Huang.
\newblock A comprehensive capability analysis of gpt-3 and gpt-3.5 series models.
\newblock {\em ArXiv}, abs/2303.10420, 2023.

\bibitem{yuan2024multifaceted}
Yefeng Yuan, Yuhong Liu, and Liang Cheng.
\newblock A multi-faceted evaluation framework for assessing synthetic data generated by large language models.
\newblock {\em arXiv preprint 2404.14445}, 2024.

\bibitem{ctabgan}
Zilong Zhao, Aditya Kunar, Robert Birke, and Lydia~Y Chen.
\newblock Ctab-gan: Effective table data synthesizing.
\newblock {\em Asian Conference on Machine Learning (PMLR)}, 2021.

\end{thebibliography}
